%% file: main.tex
\documentclass[letterpaper, 10pt, conference]{ieeeconf}  
\pdfoutput=1

\IEEEoverridecommandlockouts                              

\usepackage{xcolor}
\usepackage{amsmath,bm,times}
\usepackage{graphicx}
\usepackage{booktabs}
\usepackage{cite}
\usepackage{url}
\usepackage{amsmath,amssymb,amsfonts}
\usepackage{algorithmic}
\usepackage{graphicx}
\usepackage{textcomp}
\usepackage{tabularx}
\usepackage{subcaption}
\usepackage{flushend}

\usepackage{multirow}

\usepackage{hyperref}
\hypersetup{
    colorlinks=true,
    linkcolor=blue,
    citecolor=blue,
    filecolor=magenta,
    urlcolor=cyan}

\usepackage{epsf} 

\usepackage{xspace}

\newif\ifshowcomments
\showcommentstrue
\showcommentsfalse 
\ifshowcomments
    \usepackage[textsize=scriptsize]{todonotes}
    \newcommand{\fix}[1]{{\color{red} #1}}
    \newcommand{\todoinas}[1]{\textcolor{orange!80}{[AS: #1]}\xspace}
    \newcommand{\todoinmz}[1]{\textcolor{blue!80}{[MZ: #1]}\xspace}
\else
    \usepackage[disable,textsize=scriptsize]{todonotes}
    \newcommand{\fix}[1]{}
    \newcommand{\todoinas}[1]{}
    \newcommand{\todoinmz}[1]{}
\fi

\newcommand{\todomz}[1]{\todo[fancyline,color=blue!40]{MZ: #1}\xspace}

\newcommand{\todomw}[1]{\todo[fancyline,color=green!40]{MW: #1}\xspace}
\newcommand{\todoinmw}[1]{\todo[inline,color=green!40]{MW: #1}\xspace}

\newcommand{\OUT}[1]{}

\newcommand{\alg}{IteraPlan\xspace}

\makeatletter
\let\NAT@parse\undefined
\makeatother
\usepackage[square, numbers, sort&compress]{natbib}

\title{\LARGE \bf From Vague Instructions to Task Plans:\\A Feedback-Driven HRC Task Planning Framework based on LLMs}

\author{Afagh Mehri Shervedani$^{1}$, Matthew R.\ Walter$^{2}$, and Milo\v s \v Zefran$^{1}$
\thanks{$^{1}$A. Mehri Shervedani and Milo\v s \v Zefran are with the Robotics Laboratory, Department of Electrical and Computer Engineering, University of Illinois Chicago, Chicago, IL 60607 USA.}%
\thanks{$^{2}$M. Walter is with the Robotics Laboratory, Toyota Technological Institute at Chicago, Chicago, IL 60637 USA.}%
\thanks{This work was supported in part by the National Science Foundation under grant ECCS-2216899.}%
}

\begin{document}

\maketitle

\begin{abstract}
Recent advances in large language models (LLMs) have demonstrated their potential as planners in human-robot collaboration (HRC) scenarios, offering a promising alternative to traditional planning methods. LLMs, which can generate structured plans by reasoning over natural language inputs, have the ability to generalize across diverse tasks and adapt to human instructions. This paper investigates the potential of LLMs to facilitate planning in the context of human-robot collaborative tasks, with a focus on their ability to reason from high-level, vague human inputs, and fine-tune plans based on real-time feedback. We propose a novel hybrid framework that combines LLMs with human feedback to create dynamic, context-aware task plans. Our work also highlights how a single, concise prompt can be used for a wide range of tasks and environments, overcoming the limitations of long, detailed structured prompts typically used in prior studies. By integrating user preferences into the planning loop, we ensure that the generated plans are not only effective but aligned with human intentions. 
\end{abstract}

\input{introduction.tex}

\input{literature.tex}

\input{methodology.tex}

\input{implementation.tex}
\input{evaluations.tex}

\input{conclusion.tex}

\bibliographystyle{IEEEtranN}
\bibliography{references}

\end{document}

%% file: introduction.tex
\section{Introduction}
\label{sec: Introduction}

Planning is a fundamental aspect of robotics, enabling autonomous agents to generate sequences of actions to achieve specific goals. Traditional planning methods in the context of human-robot collaboration (HRC) 
and assistive robots can be broadly categorized into two main types: \textbf{rule-based planners} 
and \textbf{learning-based planners}. Rule-based planners rely on predefined heuristics and symbolic representations, making them interpretable at the expense of not being able to adapt to complex or dynamic environments. 
In contrast, learning-based planners, particularly those utilizing deep reinforcement learning, learn to generate plans from experience in an adaptive manner. However, they often require extensive amounts of training data and suffer from issues such as poor generalization and a lack of interpretability.

The primary objective of \textbf{task-level planning} is to decompose complex tasks into manageable subtasks and sequence
them effectively to achieve the ultimate goal. Traditional approaches to task-level planning are \textbf{rule-based} and often employ a hierarchical model of the dependencies between actions using Hierarchical Task Networks (HTNs)~\cite{erol1993toward, erol1994htn}. 
Over time, various methods have been introduced to enhance the expressive power and adaptability of HTNs. For example, the Planning Domain Definition Language (PDDL) was integrated into HTNs to improve their ability to handle complex planning problems~\cite{nau2003shop2}. Work by \citet{nooraei2014real} and \citet{mohseni2015interactive} explores how robots can learn hierarchical task models from demonstrations by integrating learning from demonstration (LfD) with Hierarchical Task Networks (HTNs), allowing them to construct HTNs dynamically rather than relying solely on predefined structures.\todomw{The discussion of how these hierarchies are built seems out-of-place here}\todoinas{I aimed to give a brief overview of HTNs and then different techniques/methods like PDDL and LfD that were integrated into HTNs. I revised the text to clarify this.}

%
%

\begin{figure}[!t]
    \centering
    \captionsetup{font=footnotesize}
    \includegraphics[width=\columnwidth]{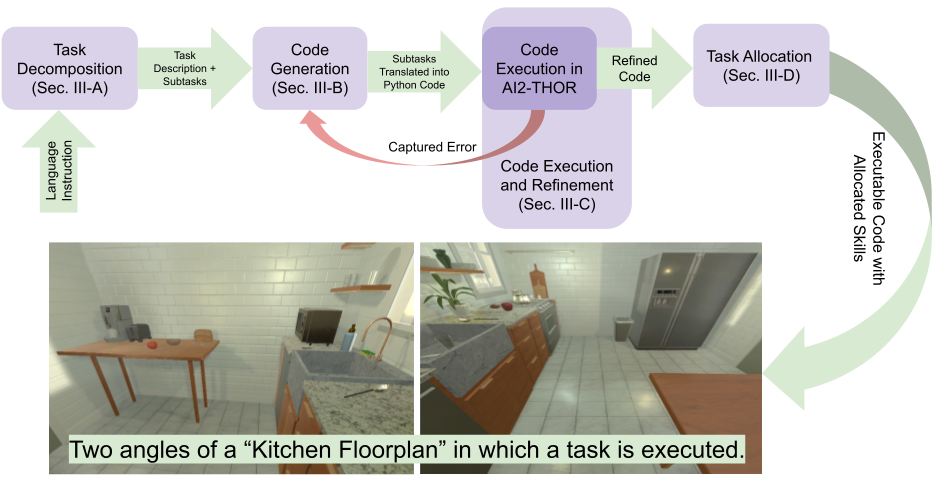}
    \caption{Overview of the \textbf{\alg} framework. The framework consists of four key components: (1) Task Decomposition, where vague, high-level instructions are transformed into structured task descriptions and subtasks; (2) Code Generation, where the subtasks are translated into executable Python code based on available skills and environmental constraints; (3) Real-Time Execution and Adaptive Code Refinement, where the generated code is executed within the AI2-THOR simulation environment, errors are detected, and iterative refinements are made to ensure feasibility; and (4) Affordance-Based Task Allocation, where executable code is adjusted to allocate actions between human and robot agents based on their capabilities.}
    \label{fig:overview}
\end{figure}
Building on progress in machine learning, researchers deployed probabilistic decision-theoretic methods such as \textbf{Markov decision processes (MDPs)}~\cite{papadimitriou1987complexity} and \textbf{reinforcement learning (RL)} that replace rule-based planners with data-driven task-level planning methods for HRC.
\citet{unhelkar2020semi} model collaborative human-robot tasks using partially observable Markov decision processes (POMDPs), specifically focusing on a human-robot handover task in a kitchen setting where a robot assists a person in preparing sandwiches. The task involves multiple steps, including fetching ingredients from cabinets, assembling them in cooking areas, and performing handovers. \citet{huang2021probabilistic} introduce a POMDP-based adaptive task planning model that accounts for both designer and operator intents in human-robot assembly. 
The study presented in~\citet{rusu2016sim} explores how RL can be used to train assistive robots to autonomously break down complex tasks (e.g., helping humans dress, fetch items, etc.) into subtasks, plan actions, and adapt to varying human needs.\todomw{Revisit for grammar}

However, while an improvement over rule-based methods, the aforementioned learning-based approaches require extensive domain-specific engineering (e.g., in defining the set of states and actions) and as a result, exhibit limited generalizability to new environments and tasks. This has led to increasing interest in more adaptive and data-driven planning approaches that take advantage of the large-scale reasoning capabilities of large language models (LLMs). Among them, SayCan~\cite{brohan2022saycan} employs an LLM (PaLM~\cite{chowdhery2023palm}) to decompose a task into a sequence of candidate subtasks that they then rank in terms of their feasibility with respect to the environment and robot capabilities using an affordance model. The model conditions its plans on the basis of both language input and real-world feasibility, enabling robots to perform assistive tasks, such as fetching objects or cleaning. Additionally, SMART-LLM~\cite{smartllm2024} uses LLMs for task-level planning in a multi-agent environment. SMART-LLM converts high-level task instructions provided as input into a multi-robot task planning problem, allowing agents to coordinate and execute tasks with high-level guidance.

Despite these advances, there remains a gap in task-level planning that seamlessly integrates human feedback, adaptive learning, and real-time interaction, particularly in complex, dynamic, and collaborative environments. Furthermore, most LLM-powered systems focus on simplifying planning through textual input without addressing the underlying nuances of human preferences and dynamic task modifications.

\textbf{Contributions of \alg:} In this paper, we address the challenges in task-level planning for HRC by proposing \textbf{\alg} (Fig. \ref{fig:overview}), an adaptive framework that leverages the strengths of LLMs, while incorporating continuous human feedback to dynamically adjust task plans. Specifically, the paper makes the following contributions:
\begin{itemize} 
    \item Unlike existing studies, where LLMs decompose \emph{explicit} high-level task instructions into subtasks, we address a more complex scenario in which the task description is \emph{implicit} in the high-level human's natural language command, such as a user stating, ``I feel hungry, what do you suggest I do?" This requires deeper reasoning and task extraction from vague or incomplete user instructions.
    \item We use a single, concise prompt for task decomposition across all types of tasks and scenes, in contrast to other LLM-based approaches that rely on long, detailed structured prompts tailored to specific tasks. This generalizability allows our framework to adapt seamlessly to a wide range of tasks and unforeseen circumstances, without requiring extensive prompt engineering.
    \item Unlike other studies where the LLM-generated plan is a fixed and non-refined output we propose a feedback-driven mechanism that fine-tunes the plan iteratively by incorporating feedback from the environment. The system adjusts the plan based on real-world interaction and accounts for user preferences.
    \item We demonstrate the effectiveness of our framework on a series of human-robot collaboration tasks, showing how it can be applied in practical assistive settings.
    \item We open-source the code and implementation of this study in: \href{https://github.com/roboticslabuic/IteraPlan.git}{https://github.com/roboticslabuic/IteraPlan.git}
\end{itemize}



%% file: literature.tex
\section{Related Work}
\label{sec: Related Work}

\subsection{Rule- and Learning-based Task-Level Planning}

\subsubsection{Rule-based Planning}
Rule-based planners rely on predefined heuristics and symbolic representations that can be more interpretable but often fail to generalize to dynamic environments. One of the earliest and widely used rule-based interaction managers was Kismet~\cite{breazeal2003toward}, which responded to social cues using predefined rules. Similarly, the Robovie robot series~\cite{robovie, robovie2} leveraged rule-based architectures to  manage interactions, utilizing episode chains to dynamically adapt behaviors based on human responses. While effective for structured interactions, these approaches were rigid and struggled with off-script scenarios.  

Hierarchical approaches, particularly HTNs~\cite{erol1993toward, erol1994htn}, were introduced to improve scalability by decomposing tasks into sequences of subtasks. SHOP~\cite{nau1999shop} and its successor SHOP2~\cite{nau2003shop2} expanded the capabilities of HTNs by incorporating the PDDL, allowing for more expressive task representations. However, despite these advances, rule-based hierarchical planners remained limited in their ability to generalize to new tasks and to dynamically adapt. Research efforts have attempted to mitigate these issues by integrating LfD into HTNs~\cite{nooraei2014real, mohseni2015interactive}, where human guidance aids robots in structuring task sequences. Extensions such as Task Insertion HTNs (TIHTNs)~\cite{xiao2017hierarchical} introduced additional flexibility, yet rule-based systems continued to struggle with adaptability and real-time learning.

\subsubsection{Learning-based Planning}
To overcome the limitations of rule-based planners, researchers have turned to probabilistic decision-theoretic approaches such as MDPs~\cite{papadimitriou1987complexity} and RL. MDPs have been particularly useful for task-level planning under uncertainty, as demonstrated by \citet{gopalan2015modeling}, where a POMDP model enabled robots to infer human intentions based upon multimodal cues. The survey by \citet{kurniawati2022partially} further highlighted the computational challenges of POMDPs, but pointed to advances in sampling-based solvers as a means to improve their practical applicability.  

Reinforcement learning has gained traction due to its ability to learn optimal policies through interaction. Deep RL methods~\cite{mnih2016asynchronous, schulman2017proximal} have demonstrated significant success in robotics, with applications extending to assistive tasks~\cite{rusu2016sim}. Inverse RL techniques, such as preference-based learning~\cite{woodworth2018preference}, have been used to infer user preferences and to personalize interactions. However, many RL-based planners require a significant number of environment interactions and  extensive training data, lack interpretability, and struggle with long-horizon task decomposition. While these methods improve adaptability compared to rule-based approaches, most existing methods do not provide a means of incorporating online human feedback in real-time, which is crucial for human-robot collaboration (HRC).  

\subsection{Large Language Models (LLMs) for Planning}
Recently, LLMs have emerged as a promising tool if not an alternative for planning, due to their ability to process high-level task descriptions and generate flexible action sequences. SayCan~\cite{brohan2022saycan} integrates an LLM (PaLM~\cite{chowdhery2023palm}) with a robotic affordance model, enabling task decomposition and execution based on real-world feasibility. Similarly, SMART-LLM~\cite{smartllm2024} employs LLMs for multi-agent task planning, allowing coordination through natural language inputs. While these approaches demonstrate the utility of LLMs for planning, they often rely on structured prompts and are unable to dynamically adjust plans based on real-time human feedback.  

In the domain of dialogue systems, Generative Pre-trained Transformers (GPT)-based planners \todomz{Define GPT.}\todoinas{Done, thanks for bringing it to my attention.} have shown potential in sequential decision-making tasks. GPT-Critic~\cite{jang2022gptcritic} utilizes offline RL to fine-tune LLMs for dialogue planning, improving response generation through iterative feedback. LSTDial~\cite{ye2024lstdial} extends this idea by incorporating short- and long-term feedback mechanisms. While these methods enhance adaptability in textual planning, their applicability to multimodal human-robot interaction remains limited. Unlike traditional planners, LLMs struggle with ambiguity in user input and require mechanisms to refine their outputs based on continuous interaction.  

\subsection{Bridging the Gap: Adaptive LLM-based Planning with Human Feedback}
Despite advances in rule-based, learning-based, and LLM-based planning, existing approaches face challenges in generalization, interpretability, and adaptability. Rule-based systems are rigid, learning-based planners require extensive data, and LLMs often generate plans in a one-shot manner without real-time refinement. OUr  proposed approach addresses these limitations by leveraging LLMs for task-level planning while integrating continuous human feedback. Unlike existing LLM-based planners, our method dynamically refines plans based on environmental interactions, enabling adaptive execution in complex, collaborative scenarios. The proposed system addresses \todomz{Is this future work, or are you doing it in this paper?}\todoinas{Done in this paper. I corrected the wording to avoid confusion.} challenges such as implicit task extraction, real-time user preference integration, and iterative plan refinement, ultimately enhancing human-robot collaboration in assistive settings.

%% file: methodology.tex
\section{Methodology}
\label{sec: Methodology}

In this section, we describe our approach to task-level planning for HRC using LLMs. \textbf{\alg}, as shown in Fig. \ref{fig:overview}, consists of four main components: (1) Task Decomposition from Vague Instructions, (2) Translation of Task Plans into Executable Code, (3) Real-Time Execution and Adaptive Code Refinement, and (4) Affordance-Based Task Allocation.

\subsection{Task Decomposition from Vague Instructions}
\label{subsec: comp1}

The first component of our approach involves transforming vague, high-level instructions into actionable task plans. Unlike previous methods, which typically start with well-defined, feasible high-level instructions and detailed environmental information, we aim to work with more ambiguous inputs. This introduces two primary challenges: first, ensuring that the task description generated based on a vague instruction is feasible in the given environment; and second, preventing the inclusion of unnecessary or redundant information about the environment in the planning process.

To address these challenges, we break the task decomposition process into two stages:

\paragraph{Stage 1: Initial Task Description Generation}
In the first stage, we provide the LLM with the vague instruction and a basic list of relevant locations and objects in the environment. The LLM uses this information to generate an initial \textbf{Task Description}---a high-level outline of the task---and a corresponding set of \textbf{subtasks} that detail the steps required to complete the task. At this stage, the model may generate a task description that is not fully compatible with the environment due to the potential vagueness of the instruction that lacks specific guidance regarding the feasibility of the plan.

\paragraph{Stage 2: Refinement and Feasibility Check}
To ensure that the generated plan is both feasible and relevant, the second stage involves refining the prompt. Using the initial Task Description and its associated subtasks from Stage~1, we update the prompt to include only the relevant environmental details: specifically, the objects and locations directly involved in the execution of the subtasks. This ensures that unnecessary details are excluded, minimizing the risk of confusion and the inclusion of irrelevant information. The LLM then refines the Task Description and subtasks based on the updated prompt, aligning the plan with the constraints of the environment and ensuring the feasibility of the task.

By decomposing the task decomposition into these two stages, we ensure that the LLM generates a practical and context-aware task plan while focusing on the elements of the environment that are essential to the task. This approach allows for effective handling of vague instructions without overwhelming the planning process with excessive information, an approach that has proven effective in other domains~\cite{fang2024transcrib3d}.

\subsection{Translation of Task Plans into Executable Code}
\label{subsec: comp2}

The second component of our methodology focuses on translating the task plan and subtasks generated in the previous component into executable Python code. This step is crucial in bridging the gap between high-level natural language planning and real-world execution within the experimental environment.

To achieve this, the LLM receives:
\begin{itemize}
    \item The refined \textbf{Task Description} and \textbf{subtasks} from the previous component.
    \item A structured prompt that contains the set of \textbf{available skills} (the term \textit{Skill} will be explained in Section \ref{subsec: action-skill}) within the experiment environment.
    \item The \textbf{properties of objects and locations} involved in the task, including constraints on feasible actions.
\end{itemize}

Using this information, the LLM generates a Python script that implements the planned sequence of \textit{skills}. However, since automatic code generation may not always result in an immediately executable or optimal script, we incorporate \textbf{human feedback} into this process.

\OUT{
\paragraph{Human Feedback and Code Refinement}
After receiving the initially generated Python code, a human collaborator evaluates its correctness, feasibility, and alignment with task objectives. If the script requires modifications—such as adjustments for unforeseen constraints or additional instructions for task execution—the human provides feedback. This feedback includes:
\begin{itemize}
    \item Specifying user preferences regarding execution details.
    \item Providing additional instructions if necessary to ensure proper execution.
    \item Highlighting errors or inconsistencies in the generated code.
\end{itemize}
}

\todoinmw{Previously, we emphasized the importance of being able to accept real-time human feedback. This text is a reference to some of that feedback, correct? If so, we might want to make this clear.\\

Meanwhile, I suspect that reviewers might have concerns with this aspect, particularly with regard to its usefulness for non-exper users (i.e., people who can not validate Pyhon code), which relates to Milos' comments below.}
After receiving the initially generated Python code, a human collaborator evaluates its correctness, feasibility, and alignment with task objectives. The code is presented in a terminal window, where the user can review it before execution. Since the generated code consists of a sequence of predefined skills, the user ensures that all referenced skills exist in the simulation environment and that objects used in the script are present in the scene. Additionally, because the system operates with vague high-level instructions, plan correctness is often subjective. The user can inject personal preferences, modifying the plan to better align with their needs.

To facilitate user interaction, the system prints the generated code in the terminal, followed by a prompt asking whether the script appears correct. If the user identifies missing elements, or incorrect assumptions, or wishes to customize the plan (e.g., modifying task details such as ingredient choices in a cooking scenario), the system requests specific feedback and iteratively refines the code based on user input. The process of fine-tuning continues until the human confirms that the generated code meets all requirements.

Importantly, \textbf{no programming expertise} is required. The generated code consists of intuitive, human-readable skill sequences, where each skill is associated with specific objects. For example, the command \texttt{SliceObject('Apple', 'Knife')} clearly indicates that the slicing skill is applied to an apple using a knife. This structured representation ensures that users can easily understand and modify the code without needing prior knowledge of Python or programming.

\todomz{This is too high level. How do you present the code to the user? Is the user supposed to debug the code to find errors and inconsistencies? How do you make this user-friendly?}\todoinas{Thanks for the comment. I revised the above to include more details.}


By integrating human feedback into the code generation process, this component ensures that the final execution aligns with both environmental constraints and user-specific preferences, making the system more adaptable and reliable in real-world human-robot collaboration scenarios.

\subsection{Real-Time Execution and Adaptive Code Refinement}
\label{subsec: comp3}

\todomz{To me this section also looks too high level, more details are needed.}

The third component of our methodology focuses on executing the generated code in the simulation environment using an agent. This agent is assumed to be \textbf{a perfect agent}, meaning it can perform all available skills within the environment flawlessly. However, 
unforeseen circumstances, such as environmental physics or unmodeled constraints, may still cause execution failures. To ensure robust execution under such conditions, we introduce a mechanism for \textbf{real-time task execution and adaptive code refinement}.

The initially generated Python code is executed in the simulation environment, and any failures that occur due to unforeseen conditions are captured. These failures could arise due to:
\begin{itemize}
    \item Environmental constraints that were not explicitly encoded in the task description.
    \item Physical interactions or dependencies that affect the feasibility of an action.
\end{itemize}

Whenever execution fails (e.g., due to collision constraints, or invalid skill execution), the system extracts error messages directly from the simulation environment’s execution logs. These messages, which may include exception traces or structured failure reports, are parsed and converted into a prompt. This prompt is then provided to the LLM as feedback, guiding it to refine the task plan iteratively. The LLM adjusts the code by incorporating new constraints, modifying execution sequences, or selecting alternative skills to ensure successful task completion, an interactive approach that has proven effective as a means to improve the reasoning capabilities of LLMs~\cite{yao2023react,fang2024transcrib3d}.\todomz{How is this done?}\todoinas{Revised.}. 

The LLM refines the code by addressing the specific failure while ensuring that subsequent steps in the task plan remain valid. This refinement process is iterative: each modified version of the code is re-executed, errors are re-evaluated, and further refinements are made until the code executes successfully. This dynamic adaptation ensures robust task execution without requiring predefined, rigid plans.
\todomz{I'm confused. You execute the code before the task allocation? Is the code executed in the real world?}\todoinas{The code is executed by one single agent in the simulator.}

\todomz{It seems you need to take Fig. 1 and break it down to explain how different pieces of your framework work.}\todoinas{Revised Fig. 1}

By integrating real-time execution monitoring with iterative refinement, this component ensures that the framework remains robust against unpredictable environmental conditions. The ability to dynamically correct and adjust execution plans enhances the overall reliability of task completion in human-robot collaboration scenarios.

\subsection{Affordance-Based Task Allocation}
\label{subsec: comp4}

The final component of our methodology focuses on the collaborative execution of tasks by both the human and the robot agents. To achieve seamless task distribution, we introduce an \textbf{affordance-based task allocation} mechanism that assigns skills based on the capabilities of each collaborator.

Each \textit{skill} is a sequence of \textit{atomic actions}, where each atomic action is associated with an affordance value that indicates which agent can perform the action based on their capabilities. The Task Allocation Component assigns skills to agents using these affordances as follows:  

\begin{itemize}  
    \item Extract the skills involved in the plan from the Python code generated in the previous stage. 
    \item Compute affordance values for each skill based on the affordances of its atomic actions using rule-based heuristics.  
    \item Generate a prompt containing the skills and their affordances and employ an LLM solely to revise the Python code, ensuring that each skill is executed by the agent best suited to the skill according to their affordance. The LLM does not reason about affordances, but instead only modifies the syntax to match the assignment.  
\end{itemize}  


The system ensures: 1) Each skill is assigned to the agent (human or robot) best suited for efficient execution. 2) Task sequencing is preserved, maintaining a coherent and goal-driven collaboration. 3) Execution remains smooth, with both agents contributing in a synchronized manner.

The affordance-based task allocation strategy enables \textbf{adaptive role distribution}, allowing the framework to dynamically adjust assignments based on environmental constraints, agent availability, or task complexity. These factors are incorporated into the heuristics used for affordance calculations, ensuring flexibility in execution.  



%% file: implementation.tex
\section{Implementation}
\label{sec: Implementation}

In this section, we describe the technical aspects of our framework, including the LLM that we used, the experimental environment, and the set of available actions.

\subsection{Large Language Model (LLM)}
Our framework is compatible with different versions of GPT, such as GPT-4o, GPT-4, and GPT-3.5, ensuring broad applicability across different LLM architectures. The model interacts with the human collaborator for iterative feedback and fine-tuning to improve task execution.

\subsection{Experimental Environment}
The experiment was conducted in the AI2-THOR simulation environment~\cite{ai2thor}, a photorealistic interactive framework for embodied AI research. AI2-THOR provides a diverse set of household environments where agents can interact with objects, navigate spaces, and perform various manipulation tasks. To align AI2-THOR's action space with real-world execution, modifications were made to the available set of actions, ensuring that the generated plans could be executed both in simulation and in real-world scenarios.

\subsection{Action Space and Skill Definition}
\label{subsec: action-skill}

The action space within the AI2-THOR environment was modified to enhance compatibility with real-world execution. Actions were selected based on their feasibility in both simulated and physical settings. Each object and location in the environment has specific affordances that define the possible interactions an agent (robot or human) can perform. To structure these interactions, we define a set of atomic actions as fundamental operations that an agent can execute. These atomic actions serve as building blocks for higher-level skills, which are structured sequences of actions enabling the execution of more complex, goal-oriented behaviors.



We formally define the set of atomic actions and their composition into structured skills. The available actions in the environment are presented in Table \ref{table:thor-actions}. Each of those actions represents a fundamental capability that can be combined into higher-level skills. Skills are defined as sequences of atomic actions, enabling structured execution of complex tasks. Table \ref{table:thor-skills} gives a summary of defined skills.

As mentioned above, we modify the set of atomic actions in the simulation environment as described in Table \ref{table:thor-actions} and introduce the set of skills as defined in Table \ref{table:thor-skills} since we want our framework to be compatible with real-world executions and be applied in practical assistive settings. For example, the action \textbf{SliceObject} in the AI2-THOR environment is contextual and doesn't require any \textbf{\textit{Tool Object}}. However, we enforce using a sharp tool object for performing this action. The same holds for action \textbf{CleanObject} with enforcing usage of a \textbf{\textit{Tool Object}} (e.g., Towel, Scrub Brush) and a \textbf{\textit{Detergent Object}} (e.g., Soap, Spray Bottle).

\begin{table}[!t]
\vspace{2mm}
    \centering
    \captionsetup{font=footnotesize}
    \resizebox{\columnwidth}{!}
    {
    \begin{tabular}{ll}
        \toprule
        \textbf{Action} & \textbf{Description} \\ 
        \midrule
        \textbf{GoToObject} $\langle$objectId$\rangle$ & Navigates towards an object. \\ 
        \textbf{OpenObject} $\langle$objectId$\rangle$ & Opens an object. \\ 
        \textbf{CloseObject} $\langle$objectId$\rangle$ & Closes an object. \\ 
        \textbf{BreakObject} $\langle$objectId$\rangle$ & Breaks an object. \\ 
        \textbf{SliceObject} $\langle$objectId$\rangle$, $\langle$toolObjectId$\rangle$ &  Slices an object using a tool. \\ 
        \textbf{SwitchOn} $\langle$objectId$\rangle$ & Switches on an object. \\ 
        \textbf{SwitchOff} $\langle$objectId$\rangle$ & Switches off an object. \\ 
        \textbf{CleanObject} $\langle$objectId$\rangle$, $\langle$toolObjectId$\rangle$,  & Cleans an object using a tool and detergent. \\ 
        $\langle$canBeUsedUpDetergentId$\rangle$ &  \\
        \textbf{PickUpObject} $\langle$objectId$\rangle$ & Picks up an object.\\
        \textbf{PutObject} $\langle$receptacleObjectId$\rangle$ & Places picked-up object inside a receptacle. \\ 
        \textbf{ThrowObject} $\langle$objectId$\rangle$ & Throws an object. \\ 
        \textbf{UseUpObject} $\langle$objectId$\rangle$ & Uses parts of an object (such as Tissue or Soap). \\ 
        \bottomrule
    \end{tabular}
    }
    \caption{Set of atomic actions available in the AI2-THOR environment.}
    \label{table:thor-actions}
\end{table}

\begin{table}[!t]
    \centering
    \captionsetup{font=footnotesize}
    \resizebox{\columnwidth}{!}
    {
    \begin{tabular}{ll}
        \toprule
        \textbf{Skill} & \textbf{Sequence of Actions} \\ 
        \midrule
        \textbf{Open an Object} & GoToObject $\langle$objectId$\rangle$ $\rightarrow$ OpenObject $\langle$objectId$\rangle$ \\ 
        \textbf{Close an Object} & GoToObject $\langle$objectId$\rangle$ $\rightarrow$ CloseObject $\langle$objectId$\rangle$\\ 
        \textbf{Break an Object} & GoToObject $\langle$objectId$\rangle$ $\rightarrow$ BreakObject $\langle$objectId$\rangle$\\ 
        \textbf{Clean an Object} & GoToObject $\langle$toolObjectId$\rangle$ $\rightarrow$ PickupObject $\langle$toolObjectId$\rangle$ $\rightarrow$ \\
         & GoToObject $\langle$canBeUsedUpDetergentId$\rangle$ $\rightarrow$ \\
         & UseUpObject $\langle$canBeUsedUpDetergentId$\rangle$ $\rightarrow$ \\
         & GoToObject $\langle$ObjectId$\rangle$ $\rightarrow$ CleanObject $\langle$ObjectId$\rangle$ \\ 
        \textbf{Switch On an Object} & GoToObject $\langle$ObjectId$\rangle$ $\rightarrow$ SwitchOn $\langle$ObjectId$\rangle$\\ 
        \textbf{Switch Off an Object} & GoToObject $\langle$ObjectId$\rangle$ $\rightarrow$ SwitchOff $\langle$ObjectId$\rangle$\\ 
        \textbf{Put an Object in a Receptacle} & GoToObject $\langle$ObjectId$\rangle$ $\rightarrow$ PickupObject $\langle$ObjectId$\rangle$ $\rightarrow$ \\
         & GoToObject $\langle$receptacleObjectId$\rangle$ $\rightarrow$ PutObject $\langle$receptacleObjectId$\rangle$ \\ 
        \textbf{Slice an Object} & GoToObject  $\langle$toolObjectId$\rangle$ $\rightarrow$ PickupObject  $\langle$toolObjectId$\rangle$ $\rightarrow$ \\
         & GoToObject  $\langle$ObjectId$\rangle$ $\rightarrow$ SliceObject  $\langle$ObjectId$\rangle$ $\rightarrow$ \\
          & ThrowObject  $\langle$toolObjectId$\rangle$\\ 
        \bottomrule
    \end{tabular}
    }
    \caption{Set of skills defined as sequences of atomic actions.}
    \label{table:thor-skills}
\vspace{-2mm}
\end{table}




%% file: evaluations.tex
\section{Experimental Evaluations}
\label{sec: Evaluations}



\subsection{Preliminary Study}
\label{subsec: prelim-study}
To evaluate the performance of our framework, we conducted experiments in the AI2-THOR simulation environment across four different floorplan types: \textbf{Kitchen}, \textbf{Living Room}, \textbf{Bedroom}, and \textbf{Bathroom}. We performed a total of 40 tests. In \textbf{Kitchen}, we considered four different tasks (Task 1: ``I feel hungry." Task 2: ``Can you make me sunny side up egg?" Task 3: ``I feel so sleepy and tired." Task 4: ``Can you help me empty the cabinets and drawers?") across five distinct kitchen floorplans, resulting in a total of 20 tests. In the \textbf{Living Room}, we considered a single task (Task 1: ``Can you help me tidy up the room?") across 5 different living room floorplans, totaling five tests. In the \textbf{Bedroom} floorplan, we looked at a single task (Task 1: ``I feel so tired, need to relax.") across five different bedroom floorplans, totaling five tests. For the \textbf{Bathroom} floorplan, we examied two different tasks (Task 1: ``I feel so tired, need to relax." Task 2: ``Can you help with cleaning the room?") across five different bathroom floorplans, leading to a total of ten tests. Each of these tests was repeated using three different LLMs: GPT-4o, GPT-4, and GPT-3.5-turbo. This resulted in a total of 120 tests conducted across all floorplan types and models. \todomz{You need to be specific about what tasks you tested. Also, you need to include some examples of generated plans perhaps.}\todoinas{All high-level instruction were added. If there is enough spcace, I'll add some examples.}

Performance was evaluated based on floorplan-wise task planning. As described in previous sections, a single concise prompt was used for task planning across different floorplan types. Unlike other state-of-the-art approaches that utilize structured prompts with detailed descriptions and numerous examples spanning multiple floorplans, our approach relies on a single example in the prompt, expecting the LLM to generalize across all floorplans. Therefore, our evaluation focuses on how well each model adapts and performs across different floorplan types without explicit training on extensive examples.

We focused on evaluating \textbf{Component 2 (\ref{subsec: comp2})} and \textbf{Component 3 (\ref{subsec: comp3})} of the framework, as \textbf{Component 1} and \textbf{Component 4} consistently perform flawlessly and do not require further assessment. Since the core planning processes occur in Components 2 and 3, our evaluation metrics are designed to measure their effectiveness.

For \textbf{Component 2}, which involves generating the optimal executable code, we defined a constraint on the maximum number of refinements allowed, ranging from 0 to 5. We measured the \textbf{Optimal Code Rate (OCR)}—the success rate of obtaining the optimal code if 0, 1, 2, 3, 4, or 5 interactions (i.e., instances of human feedback and fine-tuning) were allowed. Table \ref{table:OCR-all} shows the K-OCR, L-OCR, Bd-OCR, and Bt-OCR values for Kitchen, Living Room, Bedroom, and Bathroom floorplans respectively. Comparing the last two rows of the table, we see that allowing even only one refinement improves the performance of Component 2 significantly. GPT-4o achieves optimal executable code faster than other models, reaching OCR of 1.00 with fewer refinements in Living Room, Bedroom, and Bathroom tasks (at 2 refinements). GPT-4 needs slightly more refinements to reach 1.00 in some cases, such as the Bedroom. GPT-3.5 lags behind, requiring four refinements in the Kitchen before achieving a 1.00 OCR, indicating its difficulty in generalization. GPT-4 performs best in the Living Room task, reaching 1.00 at just 1 refinement, while GPT-4o struggles more, requiring 2 refinements.


For \textbf{Component 3}, which involves executing tasks successfully, we again defined a constraint on the maximum number of refinements allowed, ranging from 0 to 5. Since our framework employs adaptive code refinement and real-time task execution, all tasks eventually succeed. However, we analyzed how many interactions (i.e., feedback and fine-tuning cycles) were necessary to reach this guaranteed success. We measured the \textbf{Successful Execution Rate (SER)}—the rate of successful executions if 0, 1, 2, 3, 4, or 5 interactions were allowed. Table \ref{table:SER-all} shows the K-SER, L-SER, Bd-SER, Bt-SER values for Kitchen, Living Room, Bedroom, and Bathroom floorplans respectively. In most cases, GPT-4 reaches a 1.00 SER slightly earlier than GPT-4o, particularly in the Living Room (at 2 refinements). All models require more refinements (GPT-3.5, GPT-4, and GPT-4o reaching 1.00 SER at 4, 3, 5 refinements respectively) in the Kitchen. This is due to tasks being more complex in the Kitchen. Notably, the zero-shot performance (SER at 0 refinements) varies significantly 
across different GPT models. 
Overall, the table highlights the ability of LLM to adequately revise the plan if the execution fails due to unforeseen conditions of the environment.

\begin{table}[h]
    \centering
    \captionsetup{font=footnotesize}
    \resizebox{\columnwidth}{!}
    {
    \setlength{\tabcolsep}{3pt}
    \begin{tabular}{ccccccccccccc}
        \toprule
        & \multicolumn{3}{c}{\textbf{K-OCR}} & \multicolumn{3}{c}{\textbf{L-OCR}} & \multicolumn{3}{c}{\textbf{Bd-OCR}} & \multicolumn{3}{c}{\textbf{Bt-OCR}}\\
        & GPT-4o & GPT-4 & GPT-3.5 & GPT-4o & GPT-4 & GPT-3.5 & GPT-4o & GPT-4 & GPT-3.5 & GPT-4o & GPT-4 & GPT-3.5\\   \midrule   
        \textbf{5} & 1.00 & 1.00 & 1.00 & 1.00 & 1.00 & 1.00 & 1.00 & 1.00 & 1.00 & 1.00 & 1.00 &1.00\\
        \textbf{4} & 1.00 & 1.00 & \textbf{1.00} & 1.00 & 1.00 & 1.00 & 1.00 & 1.00 & 1.00 & 1.00 & 1.00 &1.00\\
        \textbf{3} & \textbf{1.00} & \textbf{1.00} & 0.95 & 1.00 & 1.00 & 1.00 & 1.00 & \textbf{1.00} & 1.00 & 1.00 & 1.00 &1.00\\
        \textbf{2} & 0.95 & 0.95 & 0.95 & \textbf{1.00} & 1.00 & 1.00 & \textbf{1.00} & 0.80 & 1.00 & \textbf{1.00} & 1.00 &\textbf{1.00}\\
        \textbf{1} & 0.85 & 0.95 & 0.75 & 0.60 & \textbf{1.00} & \textbf{1.00} & 0.60 & 0.80 & \textbf{1.00} & 0.80 & \textbf{1.00} &0.60\\
        \textbf{0} & \textbf{0.30} & \textbf{0.45} & \textbf{0.60} & \textbf{0.00} & \textbf{0.20} & \textbf{0.00} & \textbf{0.60} & \textbf{0.60} & \textbf{0.20} & \textbf{0.40} & \textbf{0.20} & \textbf{0.00}\\
        \bottomrule
    \end{tabular}}
    \caption{OCR Results for Different Floorplan Types}
    \label{table:OCR-all}
\end{table}

\begin{table}[h]
\vspace{2mm}
    \centering
    \captionsetup{font=footnotesize}
    \resizebox{\columnwidth}{!}
    {
    \setlength{\tabcolsep}{3pt}
    \begin{tabular}{ccccccccccccc}
    \toprule
    & \multicolumn{3}{c}{\textbf{K-SER}} & \multicolumn{3}{c}{\textbf{L-SER}} & \multicolumn{3}{c}{\textbf{Bd-SER}} & \multicolumn{3}{c}{\textbf{Bt-SER}}\\
    & GPT-4o & GPT-4 & GPT-3.5 & GPT-4o & GPT-4 & GPT-3.5 & GPT-4o & GPT-4 & GPT-3.5 & GPT-4o & GPT-4 & GPT-3.5\\          
    \midrule
    \textbf{5} & \textbf{1.00} & 1.00 & 1.00 & 1.00 & 1.00 & 1.00 & 1.00 & 1.00 & 1.00 & 1.00 & 1.00 &1.00\\
    \textbf{4} & 0.95 & 1.00 & \textbf{1.00} & 1.00 & 1.00 & 1.00 & 1.00 & 1.00 & 1.00 & 1.00 & 1.00 &1.00\\
    \textbf{3} & 0.95 & \textbf{1.00} & 0.95 & \textbf{1.00} & 1.00 & 1.00 & 1.00 & 1.00 & 1.00 & 1.00 & 1.00 &1.00\\
    \textbf{2} & 0.95 & 0.95 & 0.95 & 0.80 & \textbf{1.00} & 1.00 & 1.00 & 1.00 & 1.00 & 1.00 & 1.00 &1.00\\
    \textbf{1} & 0.95 & 0.90 & 0.95 & 0.60 & 0.80 & \textbf{1.00} & \textbf{1.00} & 1.00 & \textbf{1.00} & 1.00 & \textbf{1.00} &\textbf{1.00}\\
    \textbf{0} & \textbf{0.70} & \textbf{0.75} & \textbf{0.85} & \textbf{0.40} & \textbf{0.60} & \textbf{0.40} & \textbf{0.60} & \textbf{1.00} & \textbf{0.40} & \textbf{1.00} & \textbf{0.80} &\textbf{0.90}\\
    \bottomrule
    \end{tabular}}
    \caption{SER Results for Different Floorplan Types}
    \label{table:SER-all}
\vspace{-4mm}
\end{table}

\subsection{Stochasticity Study}
\label{subsec: stochastic-study}
Due to non-determinism in LLMs behavior, they produce non-deterministic responses particularly when it comes to code generation~\cite{ouyang2023llm}. To assess this, we calculate and report mean and standard deviation values for OCR and SER across different floorplans. The results summarized in Table \ref{table:OCR-SER-mean} demonstrate that when no refinements are allowed or only one refinement is allowed, the results are slightly inconsistent. That is due to the fact that we used a single concise prompt for task planning across different floorplan types. Nevertheless, allowing 2 or more refinements makes LLM responses more consistent.

\begin{table}[h]
    \centering
    \captionsetup{font=footnotesize}
    \resizebox{\columnwidth}{!}{
    \setlength{\tabcolsep}{3pt}
    \begin{tabular}{ccccccc}
        \toprule
        & \multicolumn{3}{c}{\textbf{OCR}} & \multicolumn{3}{c}{\textbf{SER}}\\
        & GPT-4o & GPT-4 & GPT-3.5 & GPT-4o & GPT-4 & GPT-3.5\\          
        \midrule
        \textbf{5} & $1.00\pm0.00$ & $1.00\pm0.00$ & $1.00\pm0.00$
                   & $1.00\pm0.00$ & $1.00\pm0.00$ & $1.00\pm0.00$\\
        \textbf{4} & $1.00\pm0.00$ & $1.00\pm0.00$ & $1.00\pm0.00$
                   & $0.98\pm0.02$ & $1.00\pm0.00$ & $1.00\pm0.00$\\
        \textbf{3} & $1.00\pm0.00$ & $1.00\pm0.00$ & $0.98\pm0.02$
                   & $0.98\pm0.02$ & $1.00\pm0.00$ & $0.98\pm0.02$\\
        \textbf{2} & $0.98\pm0.02$ & $0.93\pm0.09$ & $0.98\pm0.02$
                   & $0.93\pm0.09$ & $0.98\pm0.02$ & $0.98\pm0.02$\\
        \textbf{1} & $0.71\pm0.13$ & $0.93\pm0.09$ & $0.83\pm0.19$
                   & $0.88\pm0.19$ & $0.92\pm0.09$ & $0.98\pm0.02$\\
        \textbf{0} & $0.32\pm0.25$ & $0.35\pm0.19$ & $0.20\pm0.28$
                   & $0.67\pm0.25$ & $0.78\pm0.16$ & $0.63\pm0.27$\\
        \bottomrule
        \end{tabular}}
    \caption{OCR and SER $\textrm{Mean}\pm \textrm{STD}$ Across All Floorplans} \label{table:OCR-SER-mean}
\end{table}

\subsection{Ablation Study}
\label{subsec: ablation-study}
As we mentioned, \textbf{Component 4 (\ref{subsec: comp4})} of the framework performs flawlessly since it is rule-based. However, we are interested in evaluating its performance under a different approach: instead of using heuristics and providing pre-calculated skill affordances to the LLM, we would supply only atomic action affordances and the definition of skills. The LLM would then be responsible for reasoning, determining skill affordances, and allocating skills to the collaborators. 

We conducted this experiment using GPT-4o and analyzed the \textbf{Correctly Allocated Skills Rate (CASR)}—the ratio of correctly allocated skills to the total number of skills per floorplan. We used the output of Component 3 of the framework (the generated Python script) from our original study and asked GPT-4o to allocate skills to collaborators. As previously mentioned, 40 experiments were conducted per GPT model. For GPT-4o, a total of 282 skills were involved across these 40 experiments, distributed as follows: 187 in the Kitchen, 34 in the Living Room, 16 in the Bedroom, and 45 in the Bathroom floorplans. The results were compared to those from the rule-based task allocation in the original study, as well as a random allocation baseline where skills were assigned arbitrarily between collaborators. 

CASR results per floorplan and the overall CASR results are shown in Table \ref{table:CASR}. Out of 282 skills, 215 were correctly assigned to collaborators by the LLM, resulting in an overall CASR of 0.76. This indicates that while GPT-4o effectively handles Task Planning (Component 2) and Code Generation (Component 3), its performance in reasoning over affordances for task and skill allocation remains mediocre. In this aspect, the rule-based method from the original study outperforms the LLM-based approach.

\begin{table}[!t]
\vspace{2mm}
\centering
\captionsetup{font=footnotesize}
\resizebox{\columnwidth}{!}
{
\begin{tabular}{cccccc}
    \toprule
    &\textbf{K-CASR} & \textbf{L-CASR} & \textbf{Bd-CASR} & \textbf{Bt-CASR} & \textbf{Overall-CASR}\\          
    \midrule
    \textbf{GPT-4o} & 0.71 & 1.00 & 1.00 & 0.71 & 0.76\\
    \textbf{Random} & 0.49 & 0.58 & 0.50 & 0.51 & 0.51\\
    \textbf{Rule-based} & 1.00 & 1.00 & 1.00 & 1.00 & 1.00\\
    \bottomrule
    \end{tabular}}
    \caption{CASR Comparisons for LLM-based, Random-based, and Rule-based Task Allocation} \label{table:CASR}
\end{table}

We are interested in studying the effect of incorrect skill allocations on the overall performance of the framework. When a skill is incorrectly allocated to a collaborator, two possible scenarios arise. In the first case, the collaborator possesses the skill but executes it at a higher cost compared to another collaborator. In the second case, the collaborator does not possess the skill and is unable to execute it. The first case results in a successful but suboptimal Task Execution—while the framework achieves the intended goal, it does so with increased cost. The second case, however, leads to immediate failure, as the task cannot be completed. Regardless of whether Task Execution is suboptimal or fails outright, we are primarily interested in measuring the rate of optimal Task Execution. To quantify this, we introduce a new metric: the \textbf{Optimal Execution Rate (OER)}.


Incorrect skill allocations by GPT-4o led to 15 failed or suboptimal executions out of 40 experiments during execution. This means that 25 out of 40 instructions were executed optimally (with an OER of 0.62). In contrast, the original study had 40 optimal executions, achieving a 1.0  OER.

\begin{table}[!t]
    \centering
    \captionsetup{font=footnotesize}
    \resizebox{\columnwidth}{!}
    {
    \begin{tabular}{lccccc}
    \toprule
    &\textbf{K-OER} & \textbf{L-OER} & \textbf{Bd-OER} & \textbf{Bt-OER} & \textbf{Overall-OER}\\          
    \midrule
    \textbf{GPT-4o} & 0.40 & 1.00 & 1.00 & 0.70 & 0.62\\
    \textbf{Random} & 0.00 & 0.00 & 0.20 & 0.00 & 0.02\\
    \textbf{Rule-based} & 1.00 & 1.00 & 1.00 & 1.00 & 1.00\\
    \bottomrule
    \end{tabular}}
    \caption{OER Comparisons for LLM-based, Random-based, and Rule-based Task Allocation}\label{table:CASR}
\end{table}

%% file: conclusion.tex
\section{Conclusion}
\label{sec: Conclusion}
In this paper, we introduced a novel, adaptive framework that combines LLMs with continuous human feedback to dynamically refine task plans for human-robot collaboration (HRC). In contrast to contemporary methods, our work addresses challenging problems where humans provide natural language task specifications that are vague and for which the task objectives are implicit rather than explicit. We demonstrated that a single, concise prompt can effectively address diverse tasks and environments despite these challenges, eliminating the need for lengthy, structured prompts. By integrating human feedback directly into the planning process in an online fashion, our method is able to generate plans that are both efficient and aligned with human intentions. We evaluated our framework through various experiments conducted in the AI2-THOR simulation environment.